\pdfoutput=1

\documentclass[11pt]{article}

\usepackage{emnlp2021}
\usepackage{xr}

\usepackage{times}
\usepackage{latexsym}

\usepackage[T1]{fontenc}

\usepackage[utf8]{inputenc}

\usepackage{microtype}
\usepackage{multirow}
\usepackage{amsmath}
\usepackage{bbm}
\usepackage{amssymb}
\usepackage{graphicx}
\usepackage{enumitem}
\usepackage{float}

\title{Transformer over Pre-trained Transformer for Neural Text Segmentation with Enhanced Topic Coherence}

\author{Kelvin Lo$^{1}$~~~~Yuan Jin$^{1}$~~~~Weicong Tan$^{1}$~~~~Ming Liu$^{2}$~~~~Lan Du$^{1}$\thanks{~~Corresponding author}~~~~~Wray Buntine$^{1}$\\
$^{1}$Faculty of Information Technology, Monash University, Australia \\
{\tt \{kelvin.lo,yuan.jin,charles.tan,lan.du,wray.buntine\}@monash.edu} \\
$^{2}$School of Information Technology, Deakin University, Australia \\
{\tt m.liu@deakin.edu.au} \\}

\begin{document}
\maketitle
\begin{abstract}
This paper proposes a transformer over transformer framework, called Transformer$^2$, to perform neural text segmentation. It consists of two components: bottom-level sentence encoders using pre-trained transformers, and an upper-level transformer-based segmentation model based on the sentence embeddings. The bottom-level component transfers the pre-trained knowledge learnt from large external corpora under both single and pair-wise supervised NLP tasks to model the sentence embeddings for the documents. Given the sentence embeddings, the upper-level transformer is trained to recover the segmentation boundaries as well as the topic labels of each sentence. Equipped with a multi-task loss and the pre-trained knowledge, Transformer$^2$ can better capture the semantic coherence within the same segments. Our experiments show that (1) Transformer$^2$ manages to surpass state-of-the-art text segmentation models in terms of a commonly-used semantic coherence measure; (2) in most cases, both single and pair-wise pre-trained knowledge contribute to the model performance; (3) bottom-level sentence encoders pre-trained on specific languages yield better performance than those pre-trained on specific domains.
\end{abstract}

\section{Introduction}
Text segmentation is an NLP task that aims to break text into topically coherent segments by identifying natural boundaries of changes of topics \cite{hearst-1994-multi,moens2001generic,utiyama-isahara-2001-statistical}. It is critical in the sense that many downstream tasks can benefit from the resulting structured text, including text summarization, keyword extraction and information retrieval.

Both supervised and unsupervised learning have been applied to text segmentation. With the lack of large-quantity labels on supervised training~\cite{koshorek-etal-2018-text}, unsupervised modeling based on clustering~\cite{choi-2000-advances,chen-etal-2009-global}, Bayesian methods~\cite{du-etal-2013-topic,du2015topic,malmasi-etal-2017-unsupervised} and graph methods~\cite{glavas-etal-2016-unsupervised,malioutov-barzilay-2006-minimum} have been proposed. However, with the advancement of self-learning and transfer learning on deep neural networks, there are more recent supervised modeling approaches proposed that aim to predict labeled segment boundaries on smaller datasets.~\cite{koshorek-etal-2018-text,xing-etal-2020-improving,barrow-etal-2020-joint,Glava_Somasundaran_2020}

To the best of our knowledge, the most straightforward remedy to the above problems is knowledge transfer and distillation from pre-trained models. The rich pre-trained knowledge enables the training of a more general segmentation model on a small labeled dataset. In this paper, we propose a transformer over pre-trained transformer framework that allows different types of pre-trained information regarding sentences to be distilled to their classification for text segmentation. More specifically, the contributions of our paper are as follows:
\begin{itemize}[leftmargin=*,noitemsep, nosep]
    \item Our framework leverages pre-trained (and fixed) transformers at the bottom level to transfer (as sentence encoders) both \textit{individual} and \textit{pairwise} knowledge regarding sentences to train an upper-level transformer for segmentation.
    \item The upper-level transformer is trained with a multi-task loss with different targets, including the segment labels and the (section) topic labels.
    \item Our framework outperforms state-of-the-art segmentation models in terms of the $P_k$ metric\cite{beeferman1999statistical} across several real-world datasets in different domains and languages.
    \item A comprehensive ablation study shows that each component of our framework, in most cases, is essential by contributing to its segmentation performance.
    \item A thorough empirical study shows the impacts of language-specific and domain-specific pre-trained transformers as the sentence encoders on the segmentation performance.
\end{itemize}

\section{Related Work}
In this section, we review the past literature on the text segmentation models. These models can further be categorized into being unsupervised and supervised.

\subsection{Unsupervised Segmentation Models}
Unsupervised segmentation models are developed based on some text similarity measures. C99~\cite{choi-2000-advances}, TextTiling~\cite{hearst-1997-text} and TopicTiling~\cite{riedl-biemann-2012-topictiling} partitions texts with inter-sentence similarity matrices, lexical co-occurrence patterns and topic information from latent Dirichlet allocation (LDA)~\cite{Blei_lda_2003} respectively.
Sophisticated Bayesian models were also proposed to capture the statistical characteristics of segment (topic) generation, including topic ordering regularities \cite{du2014topic}, native language characteristics \cite{malmasi-etal-2017-unsupervised} and topic identities \cite{mota-etal-2019-beamseg}.
On the other hand, GraphSeg~\cite{glavas-etal-2016-unsupervised} and \citet{malioutov-barzilay-2006-minimum} has formulated text segmentation as graph problems.

\subsection{Supervised Segmentation Models}
Earlier supervised segmentation models \cite{galley2003discourse,hsueh2006automatic,koshorek-etal-2018-text} 
rely on heuristics-based and heavily engineered segment coherence features to train traditional classifiers (e.g. decision trees \cite{hsueh2006automatic}) that learn the relationships between the features and the segment labels. 

In recent years, deep neural network based segmentation models have started to emerge. 
A common structure for them is a two-level hierarchical network, which consist of bottom-level sentence encoder and upper-level segment boundary classifier.
Variants of LSTM~\cite{10.1162/neco.1997.9.8.1735} and Bi-LSTM are vastly used in both lower-level and upper-level models from previous studies. However, the implementations of upper-level models are more diverse among them. \citet{koshorek-etal-2018-text} and \citet{wang-etal-2018-toward} have used Bi-LSTM to predict segment boundary directly, while SECTOR~\cite{SECTOR} predicts the topic of sentence and segment boundary sequentially with LSTM. S-LSTM~\cite{barrow-etal-2020-joint} further improves the performance by incorporating the ideas of previous models. On the other hand, \citet{xing-etal-2020-improving} have introduced an auxiliary pairwise sentence coherence loss. A similar architecture is also used by \citet{lukasik-etal-2020-text}.

The closest model to ours is proposed in \cite{Glava_Somasundaran_2020}\footnote{We have been unable to compare with their model as 1) their pre-trained model has not been made public and 2) rerunning their code incurs a major run-time error irrelevant to the dataset used and the data preprocessing procedures applied.} where transformers are used for both the levels of the architecture. They also developed a semantic coherence measure on distinguishing pairs of genuine and fake text snippets as an auxiliary loss alongside the segment classification loss. However, their model does not leverage the rich and diverse knowledge extracted from pre-training tasks (e.g. masked language modeling) to encode sentences at the bottom level. Addressing this limitation, our model leverages this pre-trained knowledge for dealing with a paucity of segment labels (e.g. in specialised domains).

\section{Transformer$^2$ Architecture}

Our proposed model adopts the popular two-level network architecture for text segmentation, which consists of a lower-level sentence encoder and an upper-level segment boundary classifier.

Our model aims to enhance the learning of semantic coherence between sentences from two aspects; 1) different pre-trained embeddings, generated from different NLP tasks on large external corpora, for the same sentences can capture rich and diverse information that the target corpus does not contain; 2) sentences within same segment(i.e. sharing same topic label) tend to be semantically more coherent than those across segments (i.e. with different topic labels). The above enhancements can further improve the segmentation performance of the transformer-based classifier.

\begin{figure}[h]
\centering
\caption{\small Transformer$^{2}$ Architecture}
\includegraphics[width=0.45\textwidth]{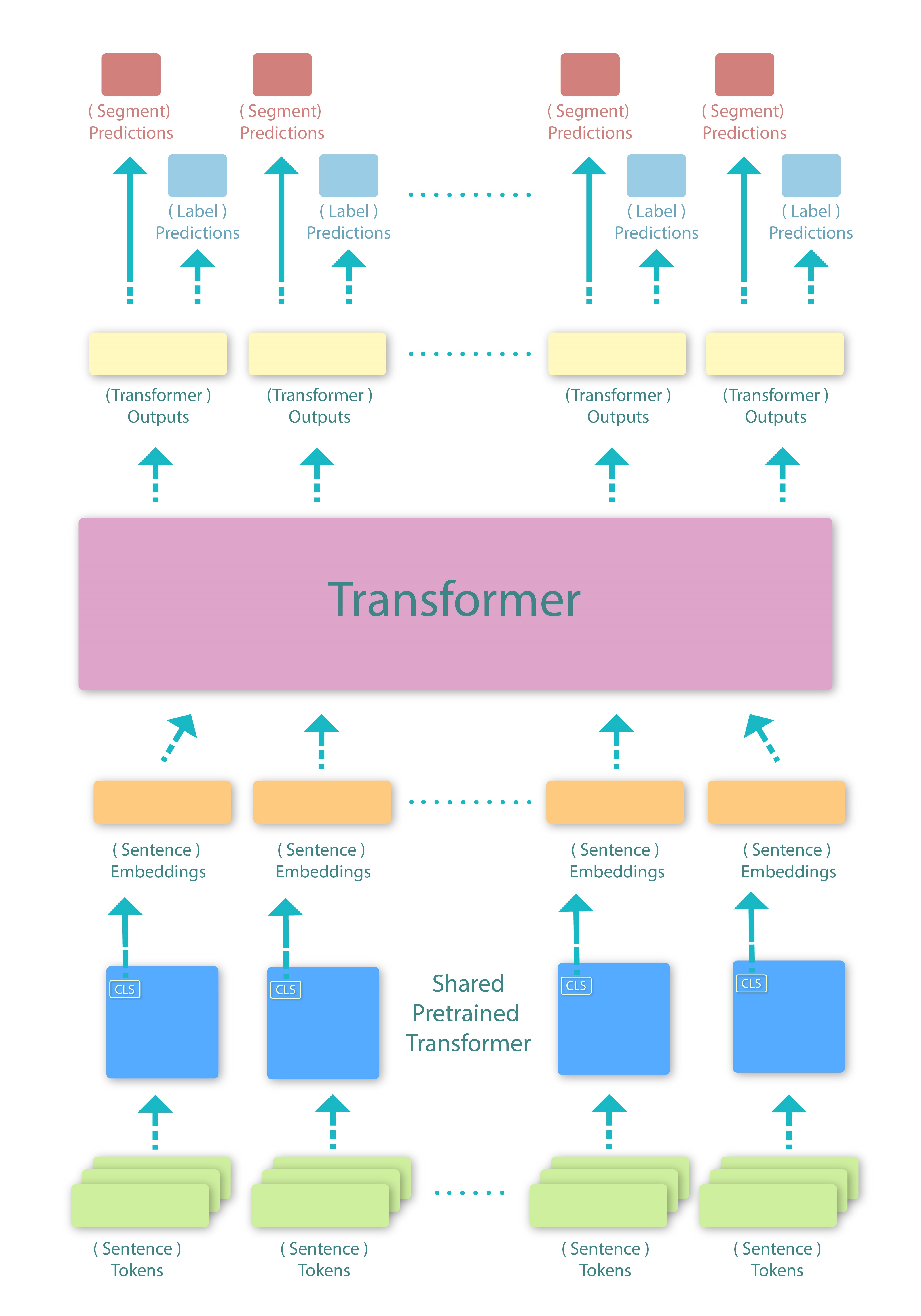}
\label{fig1}
\end{figure}

\subsection{Combining Different Pretrained Knowledge at the Bottom Level}\label{sec:bottom_level}

To introduce different prior knowledge that describes different aspects (e.g. semantics, coherence, etc.) of each sentence into the segmentation, we combine different pre-trained sentence embeddings at the bottom level. More specifically, in this paper, we concatenate the embeddings respectively generated from the [CLS] tokens with single-sentence and pairwise-sentence inputs, that is the sentence embeddings $\boldsymbol{S}:=[\boldsymbol{S}_{\text{single}};\boldsymbol{S}_{\text{pairwise}}]$. 

The single-sentence embeddings learned through masked language modelling (MLM) provide localised sentence information, while pairwise-sentence embeddings provide coherence information between consecutive sentences inherited from pairwise sentence classification tasks of pre-trained models, such as next-sentence prediction (NSP) \cite{devlin2019bert} and the sentence-order prediction (SOP) \cite{lan2019albert}. 
Further details are summarised in table \ref{tab1}.

\begin{table}[h]
\scriptsize
\centering
\caption{Transformers leveraged for the bottom level in our experiments by combining their [CLS] embedding outputs pre-trained respectively under single and pairwise tasks.}\label{tab1}
\begin{tabular}{ll|lll}
\hline
&\textbf{Single}&\multicolumn{3}{c}{\textbf{Pairwise}}\\
\textbf{Transformer} & \textbf{MLM} & \textbf{MLM} & \textbf{NSP} & \textbf{SOP}\\
\hline
BERT & \checkmark & \checkmark  & \checkmark &\\
XLNet & \checkmark & \checkmark &  &\\
RoBERTa & \checkmark&\checkmark & &\\
ALBERT & \checkmark& \checkmark & &\checkmark\\
\hline
\end{tabular}
\end{table}

\subsection{Sentence Classification at the Upper Level}

Once the sentence embeddings are obtained, we train a transformer model at the upper level of the architecture to classify 1) whether each sentence is the segment boundary and 2) the topic label of each sentence. Thus, the loss function for the upper-level transformer can be formulated as follows:
\begin{equation}
\scriptsize
\begin{split}
    L(\boldsymbol{y}_{\text{seg}}, \boldsymbol{y}_{\text{topic}}; \boldsymbol{S}, \boldsymbol{\Theta})& = L_{\text{seg}}(\boldsymbol{y}_{\text{seg}}, \hat{\boldsymbol{y}}_{\text{seg}}; \boldsymbol{S}, \boldsymbol{\Theta}) \\ & \hspace*{1em} + L_{\text{topic}}(\boldsymbol{y}_{\text{topic}}, \hat{\boldsymbol{y}}_{\text{topic}}; \boldsymbol{S}, \boldsymbol{\Theta})\\
   \hat{\boldsymbol{y}}_{\text{seg}}&:=\text{Sigmoid}(\text{Linear}_2(\text{Transformer}_{\boldsymbol{\Theta}}(\boldsymbol{S})))\\
    \hat{\boldsymbol{y}}_{\text{topic}}&:=\text{Softmax}(\text{Linear}_K(\text{Transformer}_{\boldsymbol{\Theta}}(\boldsymbol{S})))
\end{split}\label{eqn1}
\end{equation}where $\boldsymbol{S}=<\boldsymbol{s}_1,\boldsymbol{s}_2,...,\boldsymbol{s}_{I}>$, in this case, is the concatenation\footnote{With a slight abuse of notation, we reuse the symbol $\boldsymbol{S}$ from Section \ref{sec:bottom_level} to denote a sequence of all the sentences in the document.} of a sequence of embeddings of all the $\boldsymbol{I}$ sentences in the document\footnote{$\boldsymbol{I}$ denotes the maximum number of sentences in a document including the paddings.}; $\boldsymbol{y}_{\text{seg}}, \boldsymbol{y}_{\text{topic}}$ are the binary segmentation and $K$ topic labels for each sentence, while $\hat{\boldsymbol{y}}_{\text{seg}}, \hat{\boldsymbol{y}}_{\text{topic}}$ are their respective predictions. Correspondingly, linear layers with the respective output dimensions are put on top of the transformer with parameters $\boldsymbol{\Theta}$. The term $L_{\text{topic}}$ denotes an auxiliary loss on the topic labels of each sentence. Minimizing this loss forces our framework to learn semantic coherence between sentences to account for their topical similarity. As for model training, the binary segmentation loss $L_{\text{seg}}$ and the topic prediction loss $L_{\text{topic}}$ are minimized respectively as the binary and categorical cross entropy losses with respect to $\boldsymbol{\Theta}$.

\section{Experimental Results}
\subsection{Datasets} 

We used the WikiSection dataset~\cite{SECTOR} to evaluate the segmentation performance of our framework. 
It contains 38,000 full-text documents with segment information from English and German Wikipedia, each divided by topics regarding diseases and cities. The details of the corpora are summarised in Table~\ref{tab2}.

\begin{table}
\scriptsize
\centering
\caption{Summary of WikiSection Dataset}\label{tab2}
\begin{tabular}{ccccc}
\hline
\textbf{Language} & \textbf{Topic} & \textbf{Abbrev.} & \textbf{\#Subtopics} & \textbf{\#Documents}\\
\hline
English & Disease & en\_disease & 27 & 3,590\\
English & City & en\_city & 30 & 19,539\\
German & Disease & de\_disease & 25 & 2,323\\
German & City & de\_city & 27 & 12,537\\
\hline
\end{tabular}
\end{table}

\subsection{Experimental Design}
In the experiments, we leveraged both the single-sentence and pairwise-sentence pre-trained knowledge from the transformers specified in Table $\ref{tab1}$ to encode sentences at the bottom level. 
We aim to study the effects of bottom-level sentence encoders with different 1) transformer models, 2) languages and 3) domains on the segmentation performance.

Table \ref{tab3} displays the details of the transformers and their configurations (i.e. languages and domains) used in the experiments. 
More specifically, we encoded the German corpora, i.e. \textbf{de\_city} and \textbf{de\_disease}, with German BERT, which is pre-trained on the German Wikipedia dump. 
Likewise, we also encoded the domain-specific corpora, i.e. \textbf{en\_disease} and \textbf{de\_disease}, with BioClinical models, pre-trained on the MIMIC III~\cite{mimiciii} medical datasets. Detailed model configurations are listed in Appendix \ref{Model Configuration}.

\begin{table}
\scriptsize
\centering
\caption{Transformer models and their configurations (i.e. languages and domains) used in our experiments}\label{tab3}
\begin{tabular}{p{0.7cm}p{0.7cm}cccc}
\hline
\textbf{Model}&\textbf{Config.}&\textbf{en\_city} & \textbf{en\_disease} & \textbf{de\_city} & \textbf{de\_disease}\\
\hline
&English&\checkmark &\checkmark &  & \\
BERT&German& & &\checkmark  &\checkmark \\
&BioClinical& &\checkmark &  &\checkmark \\
\hline
XLNet &English& \checkmark &\checkmark &  &\\
\hline
 &English& \checkmark &\checkmark &  &\\
RoBERTa &BioMed& &\checkmark &  &\checkmark\\
\hline
ALBERT &English& \checkmark &\checkmark &  &\\
\hline
\end{tabular}\label{tab3}
\end{table}

\subsection{Evaluation Metrics \& Baselines}
Aligning with previous models, we evaluated the model performance with respect to the $P_k$ metric proposed by \citet{beeferman1999statistical}. 
It is a probabilistic metric that indicates, given a pair or words with $k$ words apart, how likely will they lie in different segments. $P_k$ values closer to 0 indicate the predicted segments are closer to ground truth,
In our experiment, the value of $k$ is set to be half of the average ground-truth segment length \cite{Pevzner:2002}.

The baselines include 1) machine learning segmentation models: C99~\cite{choi-2000-advances} and TopicTiling~\cite{riedl-biemann-2012-topictiling}, and 2) state-of-the-art deep neural models: TextSeg~\cite{koshorek-etal-2018-text}, SECTOR~\cite{SECTOR} with pre-trained embeddings, S-LSTM~\cite{barrow-etal-2020-joint} and BiLSTM+BERT~\cite{xing-etal-2020-improving}. 
We followed the default hyper-parameter settings for all the models as specified in their official implementations. 

\subsection{Transformer$^2$ Settings}
\label{Model Configuration}
For all the corpora, we have fixed several hyper-parameters of Transformer$^2$. 
We have used the Adam optimiser~\cite{DBLP:journals/corr/KingmaB14} with the learning rate being 0.0001. 
The maximum input sequence length was fixed at 150 sentences, 
as more than 94\% of the documents have less than or equal to this number of sentences across the text segmentation corpora. 
Moreover, our model has 5 transformer encoder layers with 24 self-attention heads. 
Each of the encoder layers has a point-wise feed-forward layer of 1,024 dimensions. 
For the segmentation predictions, 70\% of the inner sentences were randomly masked while all the begin sentences were not masked in order to address the imbalance class problem.

\begin{table}[t]
\scriptsize
\centering
\caption{$P_k$ values of the baselines and the best variants of Transformer$^2$ for the different datasets; Bold and underscore figures indicate the best and second best results respectively.}
\begin{tabular}{p{1.6cm}cccc}
\hline
\textbf{Model}  & \textbf{en\_disease}   & \textbf{de\_disease}   & \textbf{en\_city}     & \textbf{de\_city}     \\
\hline
C99             & 37.4          & 42.7          & 36.8         & 38.3         \\
TopicTiling     & 43.4          & 45.4          & 30.5         & 41.3         \\
TextSeg         & 24.3          & 35.7          & 19.3         & 27.5         \\
SECTOR+emb       & 26.3          & 27.5          & 15.5         & 16.2         \\
S-LSTM       & 20.0          & 18.8 & \underline{9.1} & 9.5          \\
BiLSTM+BERT       & 21.1          & 28.0 & 9.3 & 11.3          \\
\hline
$\text{Transformer}^{2}_{\text{XLNet}}$            & 25.2 & - & 11.7 & - \\
$\text{Transformer}^{2}_{\text{ALBERT}}$            & 59.1  & - & 43.6 & -  \\
$\text{Transformer}^{2}_{\text{RoBERTa}}$            & 57.2 & - & 22.7 & -  \\
\hline
\small$\text{Transformer}^{2}_{\text{BERT}}$ & \textbf{18.8} & - & \underline{9.1} & -\\
$\hspace{.3cm}\text{without }\boldsymbol{S}_{\text{single}} $ & \underline{19.9} & - & \textbf{8.2} & - \\
\hline
\small$\text{Transformer}^{2}_{\text{de\_BERT}}$ & - & \textbf{16.0} & - & \underline{7.3} \\
$\hspace{.3cm}\text{without }\boldsymbol{S}_{\text{single}} $ & - & \underline{17.1} & - & \textbf{6.8} \\
\hline
\end{tabular}\label{tab4}
\end{table}

\subsection{$P_k$ results}
\textbf{Comparison with previous models\footnote{Detailed qualitative analysis can be found in Appendix \ref{Qualitative Analysis}}} Table \ref{tab4} shows the performance of the best variants of Transformer$^2$ for different datasets and that of the baseline models in terms of the $P_k$ metric. Our models Transformer$^{2}_{\text{BERT}}$ and Transformer$^{2}_{\text{de\_BERT}}$ outperforms all previous models by a notable margin in English and German corpus respectively. 

\noindent\textbf{Ablation study of model components} We have examined the effects of single and pairwise embeddings, joint modeling on topic classification and choice of lower-level sentence encoder, summarised in tables \ref{tab5} and \ref{tab6}. The results from table \ref{tab5} shows the models yield better results without the single sentence embeddings $\boldsymbol{S}_{\text{{single}}}$ on the en\_city and de\_city datasets. This suggests that combining different pre-trained knowledge does not always improve the segmentation quality.

The results also show that the segmentation quality solely based on the change in topic label prediction labels is significantly inferior than using the segmentation labels. This is because predicting the same topic label consecutively in a multi-class setting is more difficult than the same segment label consecutively in a binary-class setting. 

On the other hand, from table \ref{tab6}, we can observe that models pre-trained on corpora in specific domains, such as BioClinical BERT, do not improve text segmentation quality compared to models pre-trained on giant language-specific corpora, such as German BERT, which is accountable to the tokenization quality of such model.

\begin{table}[!t]
\scriptsize
\centering
\caption{An ablation study on the impacts of each component of the best variants of Transformer$^2$ on $P_k$}\label{tab5}
\begin{tabular}{p{1.7cm}cccc}
\hline
\textbf{model}  & \textbf{en\_disease}   & \textbf{de\_disease}   & \textbf{en\_city}     & \textbf{de\_city}     \\
\hline
\small$\text{Transformer}^{2}_{\text{BERT}}$ & \textbf{18.8} & - & 9.1 & - \\
\scriptsize$\hspace{.3cm}\text{without }\boldsymbol{S}_{\text{single}}$ & 19.9 & - & \textbf{8.2} & - \\
\scriptsize$\hspace{.3cm}\text{without }\boldsymbol{S}_{\text{pairwise}} $ & 19.2 & - & 9.1 & - \\
\scriptsize$\hspace{.3cm}\text{without }L_{\text{topic}} $ & 20.4 & - & \textbf{8.2} & - \\
\scriptsize$\hspace{.3cm}\text{without }L_{\text{seg}} $ & 25.3 & - & 41.1 & - \\
\hline
\small$\text{Transformer}^{2}_{\text{de\_BERT}}$ & - & \textbf{16.0} & - & 7.3 \\
\scriptsize$\hspace{.3cm}\text{without }\boldsymbol{S}_{\text{single}}$ & - & 17.1 & - & \textbf{6.8} \\
\scriptsize$\hspace{.3cm}\text{without }\boldsymbol{S}_{\text{pairwise}} $ & - & 18.8 & - & 9.2 \\
\scriptsize$\hspace{.3cm}\text{without }L_{\text{topic}} $ & - & 19.5 & - & 7.2 \\
\scriptsize$\hspace{.3cm}\text{without }L_{\text{seg}} $ & - & 20.2 & - & 27.5 \\
\hline
\end{tabular}
\end{table}

\begin{table}[!t]
\small
\centering
\caption{$P_k$ values of the domain-specific BERT and RoBERTa}
\begin{tabular}{ccccc}
\hline
Sentence Encoder            & \textbf{en\_disease} & \textbf{de\_disease} \\ \hline
Transformer$^{2}_{\text{BioClinical\_BERT}}$ & 21.4                 & 45.8\\

Transformer$^{2}_{\text{BioMed\_RoBERTa}}$ &   36.4              & 50.2            \\
\hline
\end{tabular}\label{tab6}
\end{table}

\subsection{Qualitative Analysis of Transformer$^2$}
\label{Qualitative Analysis}

\begin{figure}[t]
\centering
\caption{Probabilities of segment boundaries compared to the gold-standard ones (red lines on top of each graph) on one en\_disease document where Transformer$^{2}$'s predicted probabilities are more dispersed and accurate.}
\includegraphics[width=0.45\textwidth]{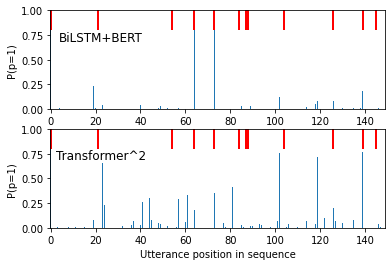}
\label{tab7}
\end{figure}

Apart from the quantitative evaluation based on the $P_k$ metric, we also conducted qualitative analysis on the segment predictions from both our model and the most competitive baseline: BiLSTM+BERT. More specifically, we randomly picked up several documents from en\_disease and de\_disease datasets, visually inspected and then summarised the difference between the segmentation styles of the best variants of Transformer${^2}$ and BiLSTM+BERT. We find that the variants of Transformer${^2}$ tend to yield \textbf{more dispersed} segment predictions across the documents, while the predictions of BiLSTM+BERT tend to be \textbf{more concentrated} and often documents are clustered as one big segment. Figure \ref{tab7} shows one such example of our finding.

\section{Conclusion and Future Work}
In this paper, we propose a transformer over pre-trained transformer framework, called Transformer$^2$, for text segmentation with a focus on enhancing the learning of the semantic coherence between sentences. The bottom level of Transformer$^2$ combines (untrainable and fixed) sentence embeddings outputted respectively from transformers pre-trained with both the single-sentence and the pairwise-sentence NLP tasks. An upper-level transformer is trained upon the combined sentence embeddings to minimize both the binary segmentation loss and the auxiliary topic prediction loss.  

The empirical results show that the best variants of Transformer$^2$ outperform several state-of-the-art segmentation models, including the deep neural models, across four real-world datasets in terms of a commonly-used segment coherence measure $P_k$. We have also conducted a comprehensive ablation study which shows that in most cases, each component of Transformer$^2$ is helpful for boosting the segmentation performance. We have also found that using language-specific pre-trained transformers at the bottom level is more useful than using domain-specific ones. For the future work, we will investigate the efficacy of Transformer$^2$ on helping the downstream NLP tasks such as text summarisation, keyword extraction and topic modelling.

\bibliography{anthology,custom}
\bibliographystyle{acl_natbib}


\end{document}